# Subgroup Analysis via Model-based Rule Forest

I-Ling Cheng
Graduate Institute of Library and
Information Science
National Chung Hsing University
Taichung, Taiwan
chengi428@gmail.com

Chan Hsu
Department of Information
Management
National Sun Yat-sen University
Kaohsiung, Taiwan
chanshsu@gmail.com

Chantung Ku
Department of Information
Management
National Sun Yat-sen University
Kaohsiung, Taiwan
kuchantung@gmail.com

Pei-Ju Lee
Department of Applied Mathematics
National Chung Hsing University
Taichung, Taiwan
pjlee@nchu.edu.tw

Yihuang Kang
Department of Information Management
National Sun Yat-sen University
Kaohsiung, Taiwan
ykang@mis.nsysu.edu.tw

*Abstract*—Machine learning models are often criticized for their black-box nature, raising concerns about their applicability in critical decision-making scenarios. Consequently, there is a growing demand for interpretable models in such contexts. In this study, we introduce Model-based Deep Rule Forests (mobDRF), an interpretable representation learning algorithm designed to extract transparent models from data. By leveraging IF-THEN rules with multi-level logic expressions, mobDRF enhances the interpretability of existing models without compromising accuracy. We apply mobDRF to identify key risk factors for cognitive decline in an elderly population, demonstrating its effectiveness in subgroup analysis and local model optimization. Our method offers a promising solution for developing trustworthy and interpretable machine learning models, particularly valuable in fields like healthcare, where understanding differential effects across patient subgroups can lead to more personalized and effective treatments.

## I. INTRODUCTION

The proliferation of big data and increased computational scalability has catalyzed rapid advances in machine learning (ML) models across diverse domains, ranging from precision medicine to personalized recommendation systems. A key driver of this progress has been representation learning techniques, notably deep neural networks (DNNs) [1, 2, 3], which learn data representations in a hierarchical, end-to-end manner, alleviating the need for handcrafted feature engineering. While remarkably expressive, these high-capacity models present a challenging black-box nature, obfuscating the mapping from inputs to outputs and giving rise to concerns surrounding interpretability, trustworthiness, fairness, and other ethical implications [4, 5].

Compounding this opacity, ML systems have been shown to be susceptible to shortcut learning [6]—learning decision rules that achieve high empirical performance on benchmark distributions yet break down under distribution shift to more complex data instances. Akin to cognitive biases in human reasoning, these learned shortcuts often yield overconfident yet fragile predictions that mask underlying flaws only revealed in real-world deployment scenarios of consequential risk (e.g., autonomous vehicle perception). Interpretable ML through techniques and applications like *subgroup analysis* plays a crucial role in developing reliable and trustworthy models, especially in domains like healthcare [7, 8, 9]. By identifying meaningful subgroups and characterizing heterogeneous patterns, subgroup analysis uncovers unique risk factors and effect modifiers associated with specific demographic or clinical populations. This knowledge enables developing equitable and personalized solutions through targeted interventions tailored to high-risk subgroups. Traditional one-size-fits-all ML and statistical models often fail to capture such nuances, leading to suboptimal performance and potential biases against vulnerable subpopulations. Interpretable subgroup discovery is therefore vital for developing ethical, trustworthy ML applications that avoid exploiting dataset biases while retaining predictive fidelity. This has motivated growing research interest in interpretable ML algorithms [4, 10, 11]. However, existing post-hoc explainability techniques like LIME [12] and SHAP [13] that approximate black-box outputs may provide misleading interpretations and fail to reliably detect errors stemming from the shortcut learning. There remains an unmet need for interpretable ML methods that elucidate faithful representations of the underlying data distributions.

We propose a novel approach, model-based Deep Rule Forests (mobDRF), which integrates rule ensemble learning and tree-based models [14] to learn interpretable, rule-encoded data representations that help subgroup discovery and analysis. Traditional rule learning algorithms often rely on simple one-level propositional logic with logic conjunctions (AND) and a few conditions, which constrains their ability to represent complex associations between targets and inputs [15]. mobDRF employs a flexible, deep model architecture that captures higher-order relationships and interactions through multi-level logic expressions. By integrating rule ensemble learning and model-based recursive partitioning [15, 16], mobDRF identifies meaningful data subgroups and learns local models that describe intricate associations.

We demonstrate mobDRF's effectiveness using real-world applications, including subgroup analysis to identify risk factors for cognitive decline in the elderly. The proposed mobDRF provides:

1. A flexible, deep model architecture that improves performance by learning intermediate rule data representations;

2. Enhanced subgroup identification and optimization of local models for better understanding of target-input associations;

3. Interpretable multi-level IF-THEN rules that correct algorithmic biases and mitigate shortcut learning problems.

The rest of this paper is organized as follows. In Section 2, we provide a brief background review related to rule learning, tree-based models, and subgroup analysis. The proposed approach, a rule representation learning algorithm with model-based Deep Rule Forests (mobDRF), is illustrated in Section 3. In Section 4, we present our experimental results with a real-world application for subgroup analysis that helps understand key risk factors of cognitive decline/impairment in the elderly population. Finally, we conclude that the proposed mobDRF can learn interpretable data representations that help improve prediction accuracy without sacrificing model interpretability.

## II. BACKGROUND AND RELATED WORK

Machine Learning (ML) algorithms have been extensively employed across various application domains to help us understand phenomena, discover patterns, and make predictions. These algorithms play crucial roles in our daily lives, from complex business decision-making to simple trip planning. Unlike traditional statistical modeling techniques, ML algorithms learn patterns from data by minimizing errors on unseen data without making assumptions about data generation [17]. Deep Learning [2], a powerful integration of artificial neural networks and Representation Learning [1] has introduced a concept of model architecture design that promotes learning data representations that better describe the associations between input features and output tasks in an automatic and layerwise manner. Additionally, the multi-modal learning [18] and multi-tasking learning [19] have further refined the architecture design by incorporating a model learning paradigm that leverages multimodal information and learns representations that may improve generalization performance in learning multiple tasks. The products of these ML algorithms have been successful in solving machine perceptual tasks, such as visual perception [20] and speech recognition [21]. However, such complex ML models have long been criticized for their black-box nature, which refers to the kinds of models that cannot provide explanations about how they arrive at predicted results without further interpretation. This issue has drawn the attention of ML communities, and researchers/practitioners have been working on interpretable ML and developing Explainable AI [4, 5, 22], which aims at eliminating algorithmic biases and promoting fair, accountable, transparent, and ethical ML applications.

In this context, interpretability in ML is defined as the capability of a model to operate within a framework that makes its workings comprehensible to humans. Our work focuses on enhancing this aspect by developing models that not only achieve high accuracy but also allow for easy interpretation through transparent rule-based representations. To this end, we propose an innovative learning algorithm that emphasizes the importance of intermediate data representations, encoded in logical rules and simple linear combinations, to balance model interpretability with performance.

Traditionally, rule learning models like Classification And Regression Trees (CART) and their derivatives/variations, such as C5.0 [23], have been favored for their straightforward, interpretable structure, which partitions the data space into subgroups (regions) defined by simple IF-THEN rules. Consider a toy example of a tree classifier for UCI ML Iris dataset [24] with two input features in Fig. 1. The tree model partitions the two-dimensional data space into five non-overlapping subgroups/regions ($R_1$, $R_2$, ... $R_5$). Each region is defined by an IF-THEN, mutually-exclusive rules with logic conjunctions of condition(s) (e.g., "IF Sepal.Length < 5.45 THEN setosa" for $R_1$).

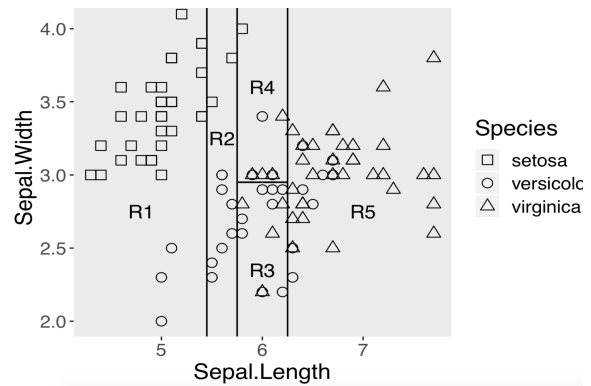

Fig. 1: A Toy Example on Learning Data Subgroups via CART

These rules can be used to *rule-encode* the original input features into a new representation/feature with rule numbers (e.g., $R_1$). The new rule-encoded feature can also be considered subgroups of interests used in further decision-making tasks. Such tree models are naturally interpretable and flexible in applications. However, they have relatively lower expressive power to learn complex patterns in data and thus usually do not perform well in terms of accuracy of prediction compared to other learning algorithms [25, 26]. Tree ensemble models, such as Random Forests [27] and XGBoost [28], were later proposed to improve such models at the cost of model interpretability. Furthermore, researchers have recently proposed deep tree ensembles, such as Deep Forests [29] and FTDRF [30], that incorporate the concepts of deep model architecture design to further improve the flexibility and capacity of the tree models. However, similar to DNNs, they are yet another black-box models. And people are still concerned about practical applications of such models when used in making high-stake decisions [4]. Here, instead, we propose to build interpretable tree ensembles that learn better intermediate data representation, a form of Deep Rule Forests (DRF) [31], which helps build interpretable models that improve both model accuracy and interpretability.

Deep Rule Forests (DRF) represent an advanced interpretative representation learning algorithm that enriches the landscape of machine learning with rule-encoded features. These are constructed by building diverse and multi-layered tree structures. Unlike conventional rule ensembles [15] that focus on merely identifying effective rule sets for prediction, DRFs are employed in tandem with other interpretable learning techniques, such as linear models, to formulate multi-level logic expressions. These expressions integrate various logical operations including conjunctions (AND), negations (NOT), and disjunctions (OR), achieving this without compromising on predictive accuracy. However, DRFs employ simple zeroth-order (propositional) logic statements to represent data layerwise. This approach often proves inadequate for regression tasks where the target is a continuous variable, as shallow regression trees struggle to capture the complex relationships between continuous targets and input features [26]. To overcome these limitations, we suggest integrating Model-Based Recursive Partitioning (MOB) trees [16] into the DRF framework. Unlike traditional CART models that predict a target based on simple averages or the most frequent category within a subgroup, MOB trees enhance the DRF process by recursively partitioning data to identify subgroups with optimally performing local models that minimize predictive errors.

Subgroup analysis [7, 8, 9] is especially critical in healthcare, where understanding the nuanced effects of treatments across different patient subgroups can significantly enhance personalized care strategies. Our mobDRF framework facilitates the identification of unique risk factors and responses to treatments within these subgroups, thereby improving both the clinical relevance and fairness of the predictive models. The versatility of MOB trees extends to a variety of applications, from analyzing patient subgroups to assessing heterogeneous treatment effects. Traditional MOB tree applications often rely on simple one-level IF-THEN rules, focusing on logic conjunctions (AND) which may not adequately address the complexities of certain datasets. In response to this challenge, our paper proposes the adoption of MOB tree ensembles within the typical DRF structure, forming a Model-Based Deep Rule Forest (mobDRF). This novel configuration leverages the strengths of both local parametric and non-parametric models, alongside multi-level logic expressions. Such a setup not only enhances the expressive power of the models but also maintains their interpretability. We will delve deeper into the specifics of the mobDRF learning process in the following sections, highlighting how it advances the field of interpretable machine learning by providing a robust framework for understanding complex, multi-dimensional data.

## III. MODEL-BASED RULE FOREST

As discussed, mobDRF is an extension of DRF that improves its ability to learn IF-THEN rules in multi-level logic expressions, as well as multiple local parametric or non-parametric models using MOB tree. This enhancement provides a better representation of the relationships between the target variable and the pre-defined explanatory features of interest. An MOB tree recursively partitions the data space with a goal to identify local models with stable parameters as well as to minimize an objective/loss function $\Psi$ that yields the estimated parameter $\theta$, as defined:

$$\hat{\theta} = \underset{\theta \in \Theta}{argmin} \sum_{i=1}^{n} \Psi(y_i, \theta)$$

where $y_i$ and $n$ are the target vector and the sample size, respectively. The MOB tree learning process can be summarized as follows:

1. Given a set of partitioning and regression input features $(x_1, x_2, ..., x_p)$ and $(z_1, z_2, ..., z_q)$, fit a parametric or non-parametric model with the regression feature and target vector $y$ in the current subgroup/subsample.

2. Assess the stability of the given parameters $\hat{\theta}$ (or cross-validation error) across each partitioning variable $x_p$.

3. Select a tree splitting feature $x_p$ and its value with highest improvement of the model fit (e.g., the greatest parameter instability).

Repeat step 1 to 3 recursively in the subsamples until pre-defined stopping criteria (e.g., the maximum depth of the tree) are met. The regression features are used to learn local models that describe the relationship between the target variable and the regressors. On the other hand, partitioning variables are involved in the tree splitting process, which divides the data space into distinct subgroups/regions based on the learned rules.

Fig. 2 illustrates the learning of new layerwise, rule-encoded features and local parametric models of mobDRF. Suppose we have a dataset consisting of $p$ partitioning and $q$ regression input features for $n$ observations, along with a target vector $y$. We initiate the process by learning multiple MOB trees of varying depths, resulting in trees with different numbers of leaf nodes. Shallow trees tend to yield oversimplified rules that fail to capture critical signals or patterns in the data, while deep trees produce complex rules that capture noise and may lead to overfitting. The tree depth (or number of leaf nodes), number of trees, and number of layers/iterations are considered hyperparameters of mobDRF. Next, we use the IF-THEN rules of an MOB tree in propositional logic to rule-encode the data, creating a new categorical feature that assigns region/leaf/subgroup numbers for a tree with $j$ leaf nodes. The original partitioning features are then replaced by the new feature set for a forest containing $m$ MOB trees. We construct a new intermediate, rule-encoded dataset with $m$ partitioning features and the original $q$ regression features to fit a new forest, which is expected to learn local parametric models tailored to specific prediction and interpretation tasks, such as measuring treatment effects. The iteration may halt when the new forest, including the learned rules and intermediate representations, no longer contributes to improving predictions (e.g., reducing error) or interpretation (e.g., generating meaningful or comprehensible rules). This approach ensures a balance between model complexity and interpretability, making it more accessible and applicable to a broader range of ML tasks.

It is important to note that mobDRF differs from typical DRF, which learns tree models by grafting tree branches in the previous layer to reduce prediction error. In contrast, mobDRF selects branches based on their ability to improve the parameter stability of the parametric or linear models, given the partitioned data. Fig. 3 demonstrates how mobDRF learns splitting rules (partitions) in a layerwise fashion.

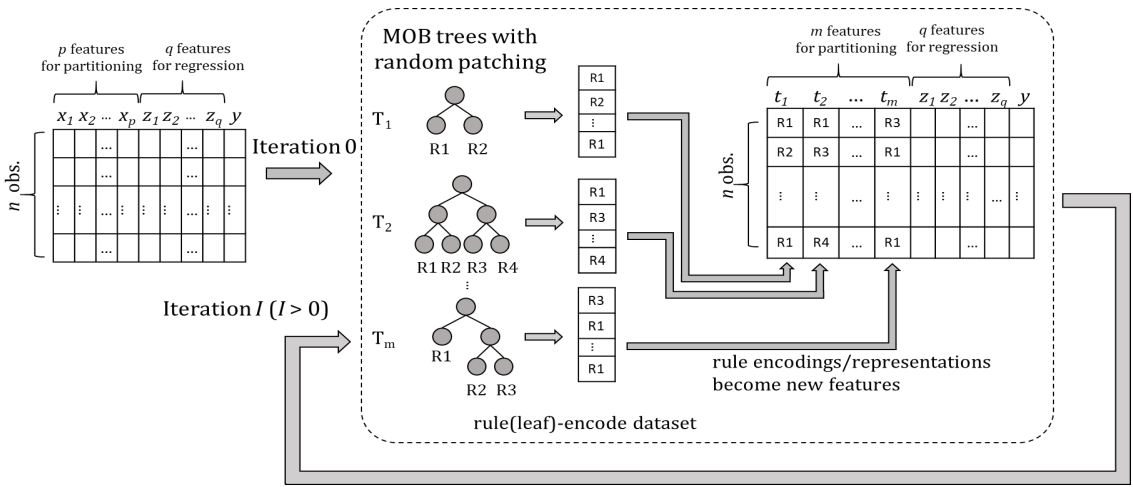

Fig 2: Model-based Deep Rule Forest Learning Process

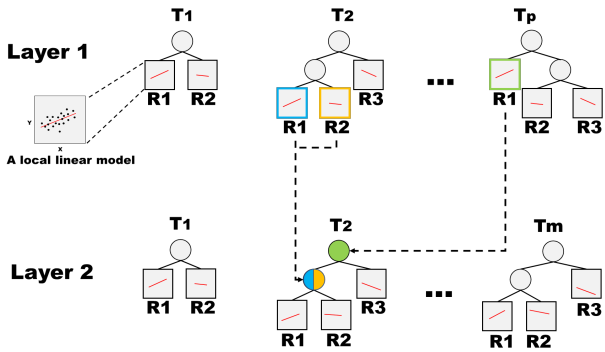

Fig 3: Layerwise Subgroup Learning with mobDRF

Consider Tree 2 ($T_2$) in Layer 2 as an example. If learning a local parametric model in Tree $p$ ($T_p$) Region 1 ($R_1$) in Layer 1 significantly improves the stability of the parameters given the target variable $y$ and regressors ($z_1, z_2, ... z_p$), $T_p R_1$ will be chosen as the first splitting feature and splitting point during the tree learning process. Subsequently, $T_2$ in Layer 2 can continue to grow by selecting appropriate branches from the previous layer (e.g., $T_2R_1$ and $T_2R_2$ in Layer 1 are selected here) until the tree reaches pre-defined growth limitations, such as tree depth and the number of leaf nodes. It is worth noting that the resampling technique, specifically random patching [32], is employed here to create multiple diverse MOB trees with various splitting criteria, ultimately facilitating the discovery of better local parametric models. In the upcoming section, we will illustrate the effectiveness of mobDRF by providing experimental results and comparing them to other established learning algorithms. This comparison will be conducted in the context of analyzing risk factors for cognitive decline in the elderly population, offering a real-world scenario to assess the performance and capabilities of mobDRF.

## IV. EXPERIMENTS AND DISCUSSION

To demonstrate our proposed method, we utilized data from the "Taiwan Longitudinal Study in Aging" (TLSA) [33]. This dataset comprises eight sections, including marriage and residence history, household schedule and social exchanges, healthcare utilization and behavior, occupational history, activity participation, residence history, economics and financial well-being, and emotional and instrumental support. The study involved re-interviewing respondents periodically from 1989 to 2015. For simplicity, we focused on the most recent data from 2015, which consists of 2,332 respondents/observations and 87 features/variables. We used 70% (n = 1,632) of the data as the training set. Our goal in using the 2015 TLSA data is to determine whether the proposed mobDRF could uncover better data representations in IF-THEN rules, ultimately helping to identify key risk factors associated with cognitive ability or decline in the elderly population. We used the sum of the Short Portable Mental Status Questionnaire (SPMSQ) [34] scores as the target feature to measure cognitive ability. In the MOB tree and mobDRF, we tentatively selected three common demographic features–age, gender (Male: 1,083, Female: 1,249), and educational level (illiteracy: 368, literacy without formal education (Literacy): 71, elementary school: 1,110, junior high: 280, senior high: 290, college and above (College): 213)–as regression features, with the remaining features serving as partitioning features.

We then fit single Prediction Rule Ensemble (PRE), Random Forest (number of trees: 500), XGBoost, MultiLayer Perceptron neural networks (MLP, 32 × 8 neurons), LASSO, Classification and Regression Trees (CART, cp = 0), and Model-Based Recursive Partitioning (MOB tree with $\alpha = 0.5$). A 3-layer mobDRF (number of trees: 500, 500, 500, max tree depth: 5, 3, 3, $\alpha = 0.1$) was created to learn intermediate rule-encoded data representation alongside LASSO, CART, and MOB trees for different applications and interpretations. All experiments were conducted on a computing server with two Intel Xeon CPUs and an NVIDIA GeForce RTX 2080 Ti GPU, and were implemented in R 4.2.3 [35] using R packages pre [15], partykit [36], xgboost [37], rpart [38], randomForest [39], and Keras [40]. Table. 1 presents the model performance evaluation in terms of training and testing Mean Absolute Error (MAE) and Root-Mean-Square Error (RMSE) on the 2015 TLSA dataset. Note that the model results on the raw TLSA 2015 (Layer 0) are denoted by dashed lines. We can see that models with relatively lower expressive power, such as CART, perform poorly on the raw data (Layer 0) compared to more complex ensemble models. However, ensemble models such as Random Forest tend to overfit the training dataset. And understanding such black-box models can be challenging without using model explainers or further interpretation. Another interesting finding is that the MOB trees, CART, and LASSO on the learned rule-encoded representations (Layer 1-3) perform comparably well, suggesting that mobDRF indeed learned data representations in rules that help improve the prediction accuracy for these models.

TABLE 1: PERFORMANCE EVALUATION on TLSA 2015

| Model | Layer | Training | | Testing | |
|---|---|---|---|---|---|
| | | MAE | RMSE | MAE | RMSE |
| PRE | - | 0.691 | 1.143 | 0.736 | 1.280 |
| Random Forest | - | 0.299 | 0.509 | 0.688 | **1.215** |
| XGBoost | - | 0.305 | 0.504 | 0.719 | 1.285 |
| MLP | - | 0.509 | 1.050 | **0.633** | 1.233 |
| LASSO | - | 0.734 | 1.181 | 0.783 | 1.293 |
| | 1 | 0.698 | 1.142 | 0.750 | 1.277 |
| | 2 | 0.719 | 1.151 | 0.775 | 1.288 |
| | 3 | 0.713 | 0.791 | 1.138 | 1.315 |
| CART | - | 0.750 | 1.226 | 0.778 | 1.313 |
| | 1 | 0.743 | 1.202 | 0.788 | 1.324 |
| | 2 | 0.671 | 1.150 | 0.731 | 1.292 |
| | 3 | 0.617 | 1.033 | 0.757 | 1.368 |
| MOB tree | - | 0.639 | 1.156 | 0.716 | 1.258 |
| | 1 | 0.619 | 1.075 | 0.680 | 1.225 |
| | 2 | 0.632 | 1.086 | <u>0.677</u> | <u>1.217</u> |
| | 3 | 0.625 | 1.077 | 0.701 | 1.252 |

Next, we would like interpretable models to help us identify subgroups as well as understand the association between the outcome and the input explanatory variables. For the 2015 TLSA data, we aim to identify key risk factors associated with cognitive ability in the elderly population. As discussed, the MOB tree can discover patient subgroups and local linear models to describe the association between the aforementioned regression and target features. The first leaf node with a rule identified by the MOB tree on the raw data (Layer 0) is as follows:

IF  iadl_score_sum ≤ 9 ∧ iadl_score_sum ≤ 3 ∧ iadl_score_sum ≤ 1 ∧ family_3 ≤ 2
THEN spmsq_score_sum = 12.1694 − 0.0726 ∗ *Female* − 0.0417 ∗ *Age* +0.3273 ∗ *Elementart_school* + 0.5994 ∗ *Junior_high* + 0.5934 ∗ *Senior_high* + 0.4786 ∗ *College* − 0.0293 ∗ *Literacy*

The IF-THEN rule suggests that, given the patient subgroup that meets the criteria "iadl_score_sum <= 1 AND family_3 <= 2" (iadl_score_sum and family_3 are "the sum of Activity of Daily Living" and "whether their family members consulted them when making decisions", respectively), the local linear model may better predict the outcome (spmsq_score_sum) and describe the relationship between cognitive ability and the three predictors (i.e., age, gender/female, and education). Compared to boolean rules with two-level or multi-level logics, such partitioning rules with simple logical conjunction (AND) have relatively lower expressive power. For example, consider the MOB tree learned from Layer 1 of mobDRF. The rule of the first leaf node (subgroup) is as follows:

IF $((T_{351} = R_1) \lor (T_{351} = R_2) \lor (T_{351} = R_3)) \land$
$((T_{401} = R_1) \lor (T_{401} = R_2) \lor (T_{401} = R_3)) \land$
$((T_{302} = R_3) \lor (T_{302} = R_6))$
THEN spmsq_score_sum = 11.7748 − 0.4021 ∗ *Female* - 0.0464 ∗ *Age* +1.1413 ∗ *Elementart_school* + 1.5376 ∗ *Junior_high* + 1.3987 ∗ *Senior_high* + 1.5751 ∗ *College* − 0.8015 ∗ *Literacy*

We can see that the logic expression in the IF statement is in CNF (AND-of-ORs) with selected tree branches and regions/rules in Layer 1. Please refer to [35] for complete variable list and descriptions. We can then expand the two-level logic expression into the original input variables for a deeper understanding of the actual conditions of the subgroup with logically equivalent statements, as follows:

IF
((iadl_3 ≤ 0 ∧ iadl_score_sum ≤ 1) ∨
 (iadl_3 > 0 ∧ iadl_6 ≤ 0))
∧
((iadl_3 ≤ 0 ∧ iadl_4 ≤ 1 ∧ lone_2 ≤ 1) ∨
 (iadl_3 ≤ 0 ∧ iadl_4 ≤ 1 ∧ lone_2 > 1 ∧ work_1 ≤ 0) ∨
 (iadl_3 ≤ 0 ∧ iadl_4 ≤ 1 ∧ lone_2 > 1 ∧ work_1 > 0))
∧
((adl_score_sum ≤ 0 ∧ ps_8 ≤ 0 ∧ household_2 > 0 ∧ social_1 ≤ 0) ∨
 (adl_score_sum > 0))
THEN spmsq_score_sum = 11.7748 − 0.4021 ∗ *Female* - 0.0464 ∗ *Age* +1.1413 ∗ *Elementart_school* + 1.5376 ∗ *Junior_high* + 1.3987 ∗ *Senior_high* + 1.5751 ∗ *College* − 0.8015 ∗ *Literacy*

Utilizing the proposed mobDRF has enabled learning more complex rules as subgroup identifications from the 2015 TLSA data, offering an interpretable data representation that facilitates understanding the association between the target outcome and input features without compromising model accuracy or interpretability. It is worth noting that trees may generate more intricate IF-THEN rules in multi-level logic when learning from the rule-encoded data representation in higher layers of mobDRF. One potential solution is to perform logic minimization to streamline the logic expressions, and/or employ mobDRF in conjunction with other learning algorithms (e.g., linear models with elastic net regularization) that can extract ranked rules, assisting in the evaluation of rule importance.

## V. CONCLUSION

We introduce a novel interpretable representation learning algorithm, model-based Deep Rule Forests (mobDRF), as a solution to the challenge of subgroup discovery and analysis in machine learning and healthcare analytics. The proposed mobDRF aims to simultaneously enhance model accuracy and interpretability by learning superior intermediate data representations in the form of human-comprehensible rules. Our experimental results demonstrate that mobDRF achieves comparable accuracy to other machine learning models while providing interpretable subgroup identifications. This approach has the potential to alleviate concerns regarding trust, safety, and ethical issues in machine learning applications. By offering interpretable models, mobDRF presents a promising solution to the accuracy-interpretability trade-off, contributing to the advancement of interpretable machine learning. It lays the groundwork for future research on inherently interpretable machine learning algorithms.